\documentclass{article}
\usepackage{spconf,amsmath,graphicx}
\usepackage{multirow}
\usepackage{url}
\usepackage{hyperref} 
\usepackage{cleveref} 
\usepackage{amsmath} 
\usepackage{breqn}
\usepackage{amssymb} 
\usepackage{tipa}
\usepackage{subfig}
\usepackage{caption}
\usepackage{graphicx}
\usepackage[font=small,skip=0pt]{caption}
\title{Multilingual analysis of intelligibility classification \\
using English, Korean, and Tamil dysarthric speech datasets}

\twoauthors
  {Eun Jung Yeo}
	{Seoul National University\\
	Department of Linguistics}
  {Sunhee Kim}
	{Seoul National University\\
	Department of \\ 
	French Language Education}
  {Minhwa Chung}
	{Seoul National University\\
	Department of Linguistics}
\begin{document}

\maketitle

\begin{abstract}

This paper analyzes dysarthric speech datasets from three languages with different prosodic systems: English, Korean, and Tamil. We inspect 39 acoustic measurements which reflect three speech dimensions including voice quality, pronunciation, and prosody. As multilingual analysis, examination on the mean values of acoustic measurements by intelligibility levels is conducted. Further, automatic intelligibility classification is performed to scrutinize the optimal feature set by languages. Analyses suggest pronunciation features, such as Percentage of Correct Consonants, Percentage of Correct Vowels, and Percentage of Correct Phonemes to be language-independent measurements. Voice quality and prosody features, however, generally present different aspects by languages. Experimental results additionally show that different speech dimension play a greater role for different languages: prosody for English, pronunciation for Korean, both prosody and pronunciation for Tamil. This paper contributes to speech pathology in that it differentiates between language-independent and language-dependent measurements in intelligibility classification for English, Korean, and Tamil dysarthric speech.

\end{abstract}


\vspace{-3mm}
\section{Introduction}



\label{sec:intro}
Dysarthria is a group of motor speech disorders resulting from a disturbance in neuromuscular control. Consequently, people with dysarthria often suffer from disturbances of respiration, phonation, articulation, resonance, and prosody  \cite{enderby2013disorders}. This results in degraded speech intelligibility, repeated communication failures, and poor quality of life. Therefore accurate, reliable, and robust intelligibility assessment is crucial for speech therapy and rehabilitation. 

Previous studies aimed to investigate acoustic indicators with high discrimination power for intelligibility classification (dysarthria severity classification). Features from diverse speech dimensions, such as features related to voice quality \cite{scholderle2016dysarthria}, pronunciation \cite{platt1980dysarthria, lansford2014vowel}, and prosody (speech rate, pitch, loudness, rhythm) \cite{workinger1991dysarthria, fox2012intensive}, were validated to be useful to intelligibility classification. Despite the promising results of previous studies, there are limitations of monolingual analysis.

Few studies addressed the acoustic analysis of intelligibility of dysarthria with a multilingual perspective. Multilingual study on Castillian Spanish, Colombian Spanish, and Czech found plosives, vowels, and fricatives to be relevant acoustic segments for dysarthria detection \cite{moro2019phonetic}. Experiments on Korean and American English pointed out smaller vowel space for dysarthric speakers compared to healthy controls \cite{kim2017cross}.  
Multilingual experiments on dysarthria detection using phonation, articulation, and prosody features reported the number of inter-word pauses per minute and median Harmonics to Noise Ratio to play significant roles for Czech and American English \cite{kovac2021multilingual}.

Whilst these studies concentrated on binary classification, multiclass automatic classification (intelligibility classification) was also performed on English and Korean using MFCCs, voice quality and prosody features \cite{hernandez2020prosody}. Further, authors pointed out the necessity of analyzing languages with different prosody systems, to better understand how supra-segmental measurements affect intelligibility, caused by different nature of languages. Experimental results suggested that different feature lists are required dependent on the language. However, the research is limited to English and Korean, which are known to be stress-timed and syllable-timed respectively.
Moreover, authors only focused on supra-segmental features and neglected phoneme pronunciation features, which are reported to greatly affect intelligibility of dysarthria \cite{moro2019phonetic, kim2017cross, yeo2021automatic}. 

This paper investigates the efficacy of diverse acoustic measurements on intelligibility classification for languages with different prosodic systems: English, Korean, and Tamil. Multilingual analyses are conducted using acoustic measurements related to voice quality, pronunciation, and prosody. The analysis is expected to help differentiate between language-independent measures and language-dependent measures for intelligibility classification of dysarthria.

The paper is organized as follows: \Cref{sec:speech_corpus} introduces speech corpora, which are TORGO database \cite{rudzicz2012torgo} for English, QoLT database \cite{choi2012dysarthric} for Korean, and SSNCE database \cite{ta2016dysarthric} for Tamil. \Cref{sec:acoustic} describes the list of acoustic measurements utilized in this study. \Cref{sec:analysis1} and \Cref{sec:analysis2} presents the two multilingual analyses, including examinations on the mean values of acoustic measurements and automatic intelligibility classification, which is followed by a conclusion in \Cref{sec: conclusion}.

\vspace{-2mm}
\section{Speech datasets}
\label{sec:speech_corpus}
\subsection{TORGO dataset}
\label{ssec:torgo}

TORGO dataset collected speech from 15 speakers, with 7 healthy speakers (4 males, 3 females) and 8 dysarthria speakers (5 males, 3 females). For dysarthric speakers, 7 speakers were diagnosed with Cerebral Palsy while 1 male speaker had amyotrophic lateral sclerosis (ALS). For severity classification, a Frenchay assessment \cite{enderby1980frenchay}, a standardized assessment tool for dysarthria, was conducted by a speech pathologist. Two speakers, one speaker, one speaker, and four speakers were graded as mild, mild-to-moderate, moderate-to-severe, and severe, respectively. Because mild and mild-to-moderate class included single speaker, we group two speakers into one moderate category.

TORGO dataset includes non-words, words, restricted sentences, and spontaneous speech. Although word materials also give useful information to intelligibility classification, sentences have greater acoustic information, especially related to prosody. Since this study aims to inspect acoustic measurements from different speech dimensions, we limit our analysis to sentences from TORGO database. Consequently, 569 utterances are used: 156 healthy utterances and 413 dysarthric utterances.
\vspace{-2mm}
\subsection{QoLT dataset}
\label{ssec:qolt}
Quality of Life Technology (QoLT) dataset is a Korean dysarthric speech corpus. The dataset recorded speech from 10 healthy speakers (5 males, 5 females) and 70 dysarthric speakers with Cerebral Palsy (45 males, 25 females). Five speech pathologists conducted the intelligibility assessment on a 5-point Likert scale, resulting in 23 mild speakers, 28 mild-to-moderate speakers, 12 moderate-to-severe speakers, and 7 severe speakers. For multilingual analysis, we combine mild-to-moderate and moderate-to-severe speakers into one category, resulting in 40 moderate speakers. All healthy speakers and 60 dysarthric speakers are from Seoul or Gyeonggi province, while 10 dysarthric speakers (9 mild, 1 moderate) are from other provinces. 

QoLT dataset consists of isolated words and restricted sentences, but we only use sentences for the experiment. The corpus contains five phonetically balanced sentences repeated twice by each speaker. Accordingly, 800 total utterances are used for the analysis: 100 utterances from healthy speakers and 700 utterances from dysarthric speakers.

\subsection{SSNCE dataset}
\label{ssec:ssnce}
SSNCE dataset is a publicly available Tamil dysarthric speech corpus. The dataset consists of 10 healthy speakers (5 males and 5 females) and 20 dysarthric speakers with cerebral palsy (13 males, 7 females). Speech intelligibility scores for dysarthric speakers are marked on a 7-point Likert scale by two speech pathologists. A score of 0 is considered healthy, 1 and 2 fall under mild, 3 and 4 under moderate, and 5 and 6 under severe. 7 mild speakers, 10 moderate speakers, and 3 severe speakers are accordingly used for the analysis. 

SSNCE dataset contains both words and sentences, but only sentences are used in this paper. The list of sentences, a total of 262 unique sentences per speaker, is a combination of common and uncommon Tamil phrases. Consequently, 7790 sentences are used in this study: 2555 healthy sentences and 5132 dysarthric sentences. 

\vspace{-2mm}
\section{acoustic measurements}
\label{sec:acoustic}
Common perceptual characteristics of speech from cerebral palsy patients include breathy, fluctuating voice \cite{scholderle2016dysarthria}, frequent articulation errors \cite{platt1980dysarthria}, reduced vowel space \cite{lansford2014vowel}, uncontrolled speech rate, monotonous pitch, reduced loudness, and irregular rhythm patterns \cite{workinger1991dysarthria, fox2012intensive}. We extract a total of 39 acoustic measurements to reflect three speech dimensions - voice quality, pronunciation, and prosody. We include features that are proven to be useful to intelligibility assessments by previous studies \cite{hernandez2020prosody, yeo2021automatic}. \Cref{sec:acoustic} introduces each acoustic measurement and describes extraction methods used in this paper.

\subsection{Voice quality features}
\label{ssec:voice_quality}
For voice quality features, relative jitter, relative shimmer, pitch perturbation quotient (PPQ), amplitude perturbation quotient (APQ), and Harmonics to Noise Ratio (HNR) are extracted. Jitter and shimmer each measures the variations of fundamental frequency (F0) and amplitude, respectively. PPQ (Period Perturbation Quotient) and APQ (Amplitude Perturbation Quotient) are also extracted as they exclude the natural change of F0 over time from jitter and shimmer. 
Lastly, HNR (Harmonics to Noise Ratio) refers to the degree of acoustic periodicity, 
which represents the energy of the periodic portion divided by the energy of the noise energy. Measurements are extracted by using the voice report function from Praat \cite{boersma2011praat}, based on Praat’s default settings. 

We also extract two voice breaks measurements: number of voice breaks and degree of voice breaks. Voice breaks are considered as inter-pulse intervals longer than 17.86ms, calculated by the distance of successive glottal pulses set to 1.25 divided by the pitch floor of 70 Hz. The degree of voice breaks is calculated by dividing the total duration of voice breaks by the total duration. Voice break measurements are also extracted by using Praat.

\subsection{Pronunciation features}
\label{ssec:pronunciation}
\subsubsection{Phoneme correctness features}
\label{sssec:phonemic}
For phoneme correctness features, 3 measurements are extracted: Percentage of Correct Consonants (PCC), Percentage of Correct Vowels (PCV), and Percentage of Correct Phonemes (PCP). PCC, PCV, and PCP each refers to the number of correctly pronounced consonants, vowels, and phonemes divided by the number of target consonants, vowels, and phonemes in an utterance. Features are extracted by using publicly released fine-tuned XLSR-53 wav2vec models for each language \cite{grosman2021xlsr53-large-english, xlsr53-korean, xlsr53-tamil}. The decoded result and the canonical phoneme sequence are aligned in phoneme units, as described in \Cref{tab:alignment}. In detail, the example sentence consists of 5 consonants (HH, L, L, R, L) and 8 vowels (IY, W, IH, AH, AW, AH, EH, AY). Comparison with the decoded results shows that 2 consonants (L, L) and 5 vowels (IY, W, AH, AW, AY) match with the canonical phoneme sequence. Hence PCC results in 2/5*100 = 40.00\%, PCV 5/8*100 = 62.50\%, and PCP 7/15*100 = 53.85\%.
\begin{table}[h]
\captionsetup{justification=centering}
\caption{Alignment example for phoneme correctness}
\label{tab:alignment}
\centering
\resizebox{0.3\textwidth}{!}{
\begin{tabular}{c}
\hline\hline
Canonical phoneme sequence \\
\hline
HH  IY  W  IH  L  AH  L  AW  AH  *  R  EH  L  AY \\
\hline
\hline
Decoded phoneme sequence \\
\hline
SH  IY  W  AO  L  AH  L  AW  AE  N  L  IY  *  AY\\
\hline
\hline
\end{tabular}
}
\end{table}
\vspace{-6mm}
\subsubsection{Vowel distortion features}
\label{sssec:vowel_space}
5 features related to vowel distortion are extracted: triangular VSA, quadrilateral VSA, Formant Centralized Ratio (FCR), Vowel Articulatory Index (VAI), and F2-ratio. VSA refers to the area of the first formant (F1) and second formant (F2) vowel diagram, determined by the corner vowels. Triangular VSA uses three corner vowels /i/, /a/, and /u/ while quadrilateral VSA uses an additional /æ/. 
FCR and VAI, proposed to solve the insensitivity of VSA measurements \cite{lansford2014vowel}, are also used as the vowel centralization indicators. Lastly, F2-ratio is the ratio of the F2 of /i/ and /u/, which are representative vowels for the front vowel and the back vowel. 
For extraction, we first create Textgrids in phoneme alignments. Montreal Forced Aligner \cite{mcauliffe2017montreal} is used for English and Korean, while time-aligned phonetic transcriptions provided by the dataset are converted into Textgrids for Tamil. Secondly, first and second formants are extracted from the center of the vowels by using Praat. 
Lastly, extracted formants are fed into the equations, as in references \cite{lansford2014vowel, yeo2021automatic}. For sentences that do not contain the required corner vowels, we interpolate the formants with the average value of a corresponding speaker.

\subsection{Prosody features}
\label{ssec:prosody}
\subsubsection{Speech rate features}
\label{sssec:speech_rate}
As for speech rate features, we include speaking rate, articulation rate, number of pauses, pause duration and phone ratio. Speaking rate and articulation rate each refers to the number of produced syllables divided by total duration of an utterance and total duration excluding pause intervals, respectively. For measurements related to pauses, intervals exceeding 0.1s of silence are counted as pauses. Phone ratio is calculated by dividing the length of the non-silent intervals by total duration. Speech rate features are extracted by using Parselmouth \cite{jadoul2018introducing}, a Python library for the Praat.
\vspace{-3mm}
\subsubsection{Pitch and Loudness features}
\label{sssec:pitch_intensity}
As for pitch and loudness features, mean, standard deviation, minimum, maximum, and range of F0 and energy are included, respectively. Intervals with 0 Hz or 0 dB are excluded from the analysis. Features are extracted by using Parselmouth.
\vspace{-3mm}
\subsubsection{Rhythm features}
\label{sssec:rhythm}
\%V, deltas, Varcos, rPVIs, nPVIs are used as representations of rhythm \cite{dellwo2003relationships}. \%V refers to the proportion of vocalic intervals. Deltas are standard deviations of vocalic or consonantal interval duration. VarcoV and VarcoC are the normalized values, where delta values are divided by the average duration of vocalic or consonantal intervals. rPVIs are the average of the duration differences between successive intervals, and nPVIs are the normalized rPVIs as \cref{eq:nPVI}.  Rhythm features are extracted by using Correlatore, a rhythmic metric analyzer \cite{mairano2010confronto}.
 \begin{align}
     &\text{nPVIs } = \frac{100}{m-1} \sum_{k=1}^{m-1}  \left| \frac{d_k - d_{k+1}}{\frac{d_k+d_{k+1}}{2}} \right| && \label{eq:nPVI}
 \end{align}

\begin{table*}[t]
\caption{Mean values of acoustic measurements }
\label{tab:analysis1}
\centering
\resizebox{0.9\textwidth}{!}{

\begin{tabular}{cc|c|c|c|c|c|c|c|c|c|c|c|c|c|c|c}

\hline
\multicolumn{2}{c|}{\multirow{2}{*}{Measurements}} & \multicolumn{5}{c|}{English} & \multicolumn{5}{c|}{Korean} & \multicolumn{5}{c}{Tamil} \\
\cline{3-17}
& & healthy & dys & mild & moderate & severe & healthy & dys & mild & moderate & severe & healthy & dys & mild & moderate & severe \\
\cline{1-17}
\multirow{7}{*}{Voice quality} & jitter & 2.31 & 2.33 & 2.37 & 1.64 & 2.79 & 1.72 & 1.83 & 1.59 & 1.95 & 1.88 & 1.47 & 1.23 & 1.23 & 1.18 & 1.37\\
& shimmer & 12.07 & 8.99 & 9.97 & 5.97 & 9.46 & 7.53 & 7.73 & 6.91 & 8.28 & 7.29 & 8.07 & 7.14 & 7.15 & 7.00 & 7.57\\
& PPQ & 1.11 & 1.14 & 1.14 & 0.70 & 1.52 & 0.78 & 0.80 & 0.68 & 0.86 & 0.83 & 0.68 & 0.59 & 0.60 & 0.54 & 0.73\\
& APQ & 6.94 & 4.94 & 5.63 & 2.85 & 5.21 & 3.25 & 3.64 & 3.16 & 3.99 & 3.21 & 3.64 & 3.48 & 3.39 & 3.34 & 4.14\\
& HNR & 9.95 & 12.94 & 11.39 & 18.03 & 11.96  & 16.27 & 16.18 & 16.82 & 15.80 & 16.31 & 14.01 & 15.63 & 15.72 & 15.67 & 15.26\\
& number of VBs & 6.83 & 9.04 & 6.86 & 9.85 & 12.84 & 6.51 & 10.87 &  9.47 & 11.06 & 14.36 & 4.05 & 3.75 &  3.73 & 3.81 & 3.61\\
& degree of VBs & 16.49 & 21.60 & 19.04 & 25.94 & 23.33 & 13.82 & 30.29 & 1.59 & 1.95 & 1.88 & 19.93  & 19.73 & 18.26 & 20.79 & 19.64 \\
\cline{1-17}
\multirow{8}{*}{Pronunciation} &  \textbf{PCC} &  \textbf{95.93} &  \textbf{80.44} & 94.25 & 71.08 & 59.76 &  \textbf{97.53} & \textbf{57.74} & 83.47 & 50.92 & 12.18 & \textbf{93.17} & \textbf{55.22} & 70.42 & 53.08 & 26.90 \\
& \textbf{PCV} & \textbf{94.51} & \textbf{77.06} & 93.16 & 64.19 & 54.58 & \textbf{99.04} & \textbf{69.44} & 88.90 & 65.96 & 25.34 & \textbf{91.52} & \textbf{52.55} &  65.90 & 50.16 & 29.35 \\
& \textbf{PCP} & \textbf{95.28} & \textbf{78.90} & 93.79 & 67.93 & 57.35 & \textbf{98.20} & \textbf{62.84} & 85.85 & 57.48 & 17.90 & \textbf{92.63} & \textbf{54.47} & 69.08 & 52.28 & 27.68 \\
\cline{2-17}
& \textbf{triangular VSA} & \textbf{109791} & \textbf{84913} & 110459 & 50541 & 60574 & \textbf{199831} & \textbf{111081} & 149994 & 97226 & 62400 & \textbf{207104} & \textbf{82208} & 97715 & 85328 &35626 \\
& \textbf{quadrilateral VSA} & \textbf{162175} & \textbf{130753} & 169389 & 102766 & 74430 & \textbf{202496} & \textbf{110192} & 149031 & 95282 & 67779 & \textbf{240516} & \textbf{95325} & 114907 & 97589 & 42085 \\
& \textbf{FCR} & \textbf{1.28} & \textbf{1.33} & 1.25 & 1.38 & 1.46 & \textbf{1.35} & \textbf{1.60} & 1.40 & 1.67 & 1.85 & \textbf{1.15} & \textbf{1.46} & 1.37 & 1.47 & 1.64\\
& \textbf{VAI} & \textbf{0.80} & \textbf{0.76} & 0.81 & 0.73 & 0.69 & \textbf{0.78} & \textbf{0.66} & 0.75 & 0.63 & 0.55 & \textbf{0.89} & \textbf{0.70} & 0.74 & 0.70 & 0.63 \\
& \textbf{F2-Ratio} & \textbf{1.36} & \textbf{1.37} & 1.43 & 1.31 & 1.29 & \textbf{1.56} & \textbf{1.34} & 1.55 & 1.25 & 1.12 & \textbf{2.13} & \textbf{1.55} & 1.67 & 1.53 & 1.31\\
\cline{1-17}
\multirow{25}{*}{Prosody} & speaking rate & 2.96 & 2.76 & 2.80 & 2.49 & 2.90 & 3.41 & 2.41 & 2.79 & 2.33 & 1.59 & 2.49 & 2.56 & 2.72 & 2.43 & 2.62\\
& \textbf{articulation rate} & \textbf{6.39} & \textbf{5.69} &  6.11 & 4.78 & 5.57 & \textbf{5.67} & \textbf{5.33} & 5.20 & 5.33 & 5.74 & \textbf{5.46} & \textbf{4.36} & 4.57 & 4.31 & 4.04\\
& \textbf{number of pauses} & \textbf{1.40} & \textbf{2.89} & 1.65 & 4.90 & 3.80 & \textbf{1.49} & \textbf{5.23} & 3.49 & 5.73 & 8.13 & \textbf{2.14} & \textbf{2.32} & 2.25 & 2.31 & 2.48 \\
& pause duration & 2.44 & 3.25 & 2.58 & 4.18 & 3.87 & 1.52 & 4.90 & 2.93 & 5.00 & 10.90 & 1.85 & 1.38 &  1.37 & 1.47 & 1.11 \\
& phone ratio & 0.46 & 0.49 & 0.46 & 0.52 & 0.52 & 0.60 & 0.47 & 0.54 & 0.45 & 0.32 & 0.46 & 0.59 & 0.69 & 0.57 & 0.66\\
\cline{2-17}
& F0 mean & 60.03 & 79.44 & 69.53 & 87.45 & 93.21 & 98.57 & 83.76 & 91.55 & 80.91 & 74.39 & 88.44 & 123.51 & 110.04 & 124.31 & 148.94\\
& \textbf{F0 std} & \textbf{90.10} & \textbf{94.77} & 96.06 & 90.70 & 95.43 & \textbf{95.64} & \textbf{99.80} & 97.76 & 99.29 & 109.38 & \textbf{106.23} & \textbf{218.21} & 95.52 & 114.46 & 112.01\\
& F0 min & 106.53 & 101.76 & 112.40 & 91.98 & 87.94 & 102.17 & 102.22 & 103.20 & 102.50 & 97.36 & 149.14 & 107.46 & 141.44 & 157.15 & 153.33\\
& F0 max & 420.77 & 410.33 & 394.67 & 401.32 & 449.69 & 261.97 & 405.76 & 339.47 & 436.04 & 450.51 & 311.67 & 151.08 & 258.36 & 307.79 & 285.48\\
& F0 range & 314.23 & 308.57 & 282.27 & 309.35 & 361.75 & 159.80 & 303.54 & 236.27 & 333.54 & 353.15 & 162.53 & 136.06 & 116.92 & 150.64 & 132.15\\
\cline{2-17}
& energy mean & 53.02 & 58.38 & 56.36 & 57.50 & 63.24 & 51.64 & 48.44 & 49.52 & 48.35 & 45.46 & 40.21 & 42.32 & 42.03 & 40.68 & 48.45\\
& energy std & 12.26 & 14.01 & 13.19 & 15.96 & 14.10 & 19.34 & 16.23 & 17.48 & 15.64 & 15.45 & 22.07 & 47.04 & 17.57 & 17.91 & 17.40\\
& \textbf{energy min} & \textbf{39.92} & \textbf{40.52} & 41.56 & 34.25 & 43.50 & \textbf{15.17} & \textbf{16.62} & 15.89 & 17.16 & 15.97 & \textbf{12.63} & \textbf{17.72} & 12.48 & 11.98 & 13.42\\
& energy max & 77.50 & 83.44 & 81.96 & 82.95 & 86.84 & 77.24 & 77.01 &  75.84 & 77.10 & 80.36 & 73.76 & 66.99 & 66.45 & 66.29 & 70.57\\
& energy range & 37.58 & 42.91 & 40.41 & 48.70 & 43.34 & 62.07 & 60.39 & 59.95 & 59.94 & 64.39 & 61.14 & 66.99 & 53.97 & 54.30 & 57.15\\
\cline{2-17}
& \textbf{\%V} & \textbf{41.72} & \textbf{43.56} & 41.45 & 45.65 & 46.18 & \textbf{84.07} & \textbf{87.46} & 85.91 & 87.70 & 91.18 & \textbf{47.09} & \textbf{55.27} &  54.78 & 54.28 & 59.68\\
& \textbf{deltaV} & \textbf{60.70} & \textbf{93.36} & 63.19 & 131.37 & 124.19 & \textbf{475.79} & \textbf{1215.04} & 787.42 & 1153.54 & 2971.46 & \textbf{62.06} & \textbf{76.47} & 71.92 & 76.44 & 87.18\\
& \textbf{deltaC} & \textbf{73.28} & \textbf{106.23} & 73.72 & 133.00 & 151.02 & \textbf{53.38} & \textbf{150.04} & 103.58 & 153.93 & 280.42 & \textbf{59.79} & \textbf{66.69} & 65.60 & 69.65 & 59.37\\
& VarcoV & 53.18 & 50.87 & 52.66 & 48.02 & 49.52 & 90.88 & 94.92 & 90.45 & 92.84 & 121.45 & 52.51 & 51.76 &  47.64 & 54.26 & 53.02\\
& \textbf{VarcoC} & \textbf{50.89} & \textbf{54.82} & 49.86 & 57.54 & 62.76 & \textbf{45.04} & \textbf{76.40} & 60.47 & 78.29 & 117.91 & \textbf{50.55} & \textbf{58.03} & 54.30 & 61.05 & 56.71\\
& \textbf{rPVIV} & \textbf{66.20} & \textbf{102.61} & 67.47 & 145.11 & 139.97 & \textbf{454.49} & \textbf{1101.09} & 758.96 & 1083.72 & 2324.55 & \textbf{73.08} & \textbf{87.19} & 81.13 & 87.78 & 99.35\\
& \textbf{rPVIC} & \textbf{81.85} & \textbf{115.28} & 77.76 & 141.92 & 170.44 & \textbf{66.88} & \textbf{174.43} & 120.21 & 179.87 & 321.56 & \textbf{72.74} & \textbf{79.73} & 78.69 & 83.05 & 71.09\\
& nPVIV & 55.85 & 54.00 & 55.22 & 52.42 & 52.79 & 84.37 & 91.71 & 86.47 & 92.23 & 105.95 & 60.27 & 58.18 & 52.27 & 61.48 & 61.00 \\
& \textbf{nPVIC} & \textbf{56.89} & \textbf{58.37} & 53.14 & 63.30 & 65.06 & \textbf{53.44} & \textbf{79.91} & 55.58 & 82.73 & 107.32 & \textbf{59.81} & \textbf{66.65} & 62.57 & 69.52 & 65.52\\
\hline

\end{tabular} 
}
\end{table*}

\vspace{-3mm}
\section{ANALYSIS 1: Multilingual Analysis}
\label{sec:analysis1}

In \Cref{sec:analysis1}, multilingual analysis is conducted by comparing the mean values of acoustic measurements between severity levels (intelligibility levels). To find measurements that show significant differences between severity levels based on intelligibility scores, Kruskal-Wallis H test is performed, because most measurements did not satisfy the Kolmogorov-Smirnov normality test. According to statistical results, every measurements show significant differences except nPVIV for English, and min F0 for Korean ($p<.05$). \Cref{tab:analysis1} presents the mean values for the extracted measurements, with dys columns indicating average values of all dysarthric speakers. Measurements with similar patterns across languages are indicated in bold.

\subsection{Voice quality features}
\label{ssec:vq_results}
For voice quality features, English speakers with dysarthria have higher values for all acoustic measurements compared to healthy speakers, except for shimmer and APQ. While Korean dysarthric speakers have higher values for all measurements but HNR, Tamil dysarthric speakers have lower values for all measurements except HNR. 

\vspace{-2mm}
\subsection{Pronunciation features}
\label{ssec:pr_results_and_vd}

Features related to phoneme correctness show that all three measurements, PCC, PCV, and PCP get constantly lower as the severity level of dysarthria gets worse in all three languages. 
Vowel distortion features also show similar patterns across languages. Smaller triangular VSA, quadrilateral VSA, higher FCR, lower VAI and F2-Ratio are found as the intelligibility level gets lower. While Korean and Tamil healthy speakers keep these tendencies compared to mild dysarthric speakers, English healthy speakers show the opposite trend. English healthy speakers have slightly smaller sizes of VSAs and higher FCR, lower VAI, and lower F2-Ratio compared to mild speakers. 

\vspace{-2mm}
\subsection{Prosody features}
\label{sssec:pro_results}
\subsubsection{Speech rate}
\label{sssec:sr}
English and Korean dysarthric speakers show slower speaking rate and articulation rate compared to healthy speakers, while Tamil dysarthric speakers have faster speaking rate and slower articulation rate. While healthy speakers produce less number of pauses than dysarthric speakers in three languages, pause duration measurements illustrate different patterns by languages. For English, longer pauses and increased phone ratio are found in dysarthric speakers. In contrast, Korean dysarthric speakers have longer pauses but smaller phone ratio. Tamil dysarthric speakers have shorter pauses but larger phone ratio, compared to healthy speakers.
\vspace{-3mm}
\subsubsection{Pitch and Loudness}
\label{sssec:pitch_int}
English dysarthric speakers have higher mean/std F0 but lower min/max/range F0 in contrast to healthy speakers. The analysis on Korean dataset presents similar min F0, lower mean F0, and higher std/max/range F0. Tamil dysarthric dataset shows higher values for all measurements except range F0.
Meanwhile, English dysarthric speakers show higher values for all intensity measurements compared to healthy speakers. However, Korean dysarthric speakers have lower values for all intensity measurements besides min energy, while Tamil dysarthric speakers produce speech with higher mean/std/min/range energy but lower max energy.
\vspace{-3mm}
\subsubsection{Rhythm}
\label{sssec:rhythm}
Dysarthric speakers generally have greater values in all rhythm measurements except for VarcoV and nPVIV in English and Tamil. English dysarthric speakers have smaller VarcoV and similar nPVIV while Tamil dysarthric speakers have both smaller VarcoV and nPVIV. Korean dysarthric speakers have larger values in all measurements.
\vspace{-2mm}
\subsection{Discussion}
\label{sssec:stats_results}
Analysis on the aspect of mean values implies while some measurements share similar patterns across languages, others show different patterns by languages. Firstly, no measurements related to \textbf{voice quality} share common pattern across English, Korean, and Tamil. 
Second, all \textbf{pronunciation} measurements show the same tendency from healthy to dysarthria, with lower values for PCC, PCV, PCP, triangular VSA, quadrilateral VSA, VAI, F2-ratio, and higher values for FCR. The result suggests speakers with dysarthria have limitations in pronouncing accurate phonemes and large enough vowel space.
Lastly, a total of 11 \textbf{prosody} measurements show similar patterns across three languages: slower articulation rate, frequent number of pauses, higher std F0, min energy, \%V, deltaV, deltaC, VarcoC, rPVIV, rPVIC, and nPVIC. Analysis on \underline{speech rate} indicates dysarthric speakers have slower articulation rate and frequent pauses. 
\underline{Pitch and loudness} measurements manifest dysarthric speakers have limitations in producing stable F0s and need greater minimum energy for speech production. Lastly, \underline{rhythm} measurements suggest dysarthric speakers have trouble in controlling durations of phonemes.

To sum up, English, Korean, and Tamil dysarthric speakers generally produce sentences with inaccurate pronunciations, slower articulation rate, frequent pauses, fluctuating F0, greater minimum energy, longer vowels, and excessive variations of phoneme duration compared to healthy speakers. On the other hand, the result also implies that each language has different patterns, especially for prosody features.

\vspace{-2mm}
\section{Analysis II: Automatic intelligibility classification}
\label{sec:analysis2}

In \Cref{sec:analysis2}, we move on to the problem of how acoustic measurements contribute to intelligibility classification. Results from automatic intelligibility classification are expected to provide insights to this problem. Experiment consists of three stages: feature extraction, feature selection, and classification. At feature extraction stage, measurements introduced in \Cref{sec:acoustic} are extracted. Descriptions related to feature selection and classification are described in \Cref{ssec:feature_selection}. 
\vspace{-2mm}
\subsection{Feature selection and Classification}
\label{ssec:feature_selection}
For feature selection, we implement Extra Trees Classifier (ETC) using scikit-learn library \cite{kramer2016scikit}. ETC discards irrelevant features for classification by computing the feature importance. Feature importance is calculated by the reduction in the likelihood of misclassification when each feature is included. ETC algorithm is fed with standardized measurements by using Standard Scaler. At classification stage, rbf support vector machine (SVM) is performed, with C and gamma grid searched between $10^{-4}$ and $10^4$. Leave-One-Subject-Out Cross Validation (LOSOCV) is used for evaluation.
\vspace{-2mm}
\subsection{Results and Discussion}
\label{ssec:Results}
\Cref{tab:classification} presents the classification results. English, Korean, and Tamil show a relative increase of 13.67\%, 6.98\%, and 3.86\% with selected features of 13, 8, and 12 measurements, respectively: (\textbf{English}) shimmer, APQ, PCC, PCV, PCP, quadrilateral VSA, VAI, mean/std/min/max/range loudness, std F0; 
(\textbf{Korean}) PCC, PCV, PCP, pause duration, phone ratio, std loudness, max F0, deltaV; 
(\textbf{Tamil}) PCC, PCV, PCP, VAI, std/max/min/range loudness, mean/std/min/max F0. 
The selected measurement lists suggest that different measurements affect intelligibility classification for different languages. On the other hand, PCC, PCV, PCP, and std loudness are commonly selected across languages. Referring to the results in \Cref{sec:analysis1}, PCC, PCV, PCP showed similar patterns between languages, but standard loudness did not share a trend across languages.

\begin{figure}[htpb]
    \centering
    \captionsetup{justification=centering}
    \includegraphics[width=0.4\textwidth ]{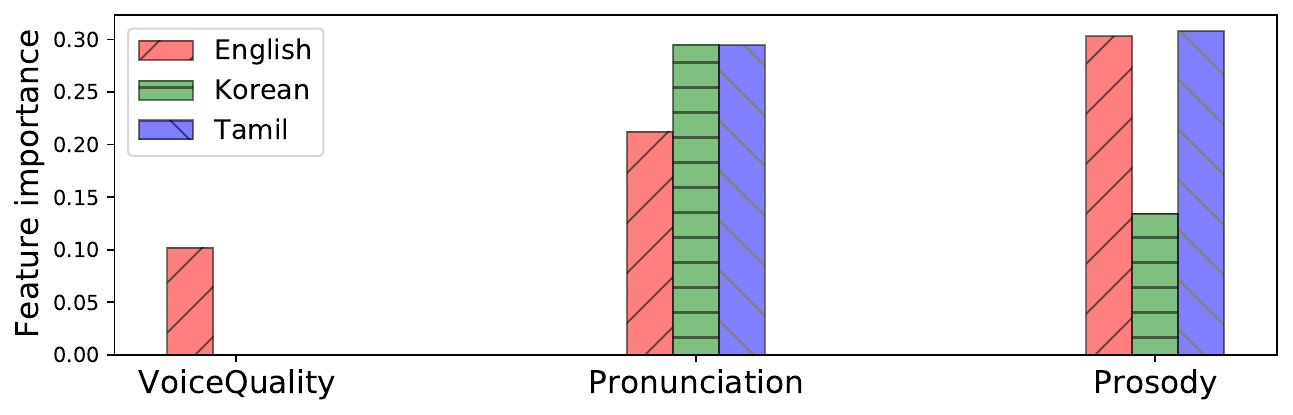}
    \caption{Feature importance}
    \label{fig:importance}
\end{figure}

Further, feature importance of selected measurements are examined. As illustrated in \Cref{fig:importance}, English majorly uses prosody information in comparison to pronunciation information for intelligibility classification. The opposite is apparent in Korean, heavily relying on pronunciation than prosody measurements, while Tamil evenly gains information from both pronunciation and prosody measurements. The result also implies different measurements play significant roles in intelligibility classification for different languages.

\begin{table}[h]
\caption{
Classification results
}
\label{tab:classification}

\centering
\resizebox{0.4\textwidth}{!}
{



\begin{tabular}{c|c|c|c}
\hline
Language & all & selected (\#F) & relative increase \\
\hline
English & 26.84 & \textbf{30.51} (13) & 13.67 \\
\hline
Korean & 71.62 & \textbf{76.62} (8) &  6.98 \\
\hline
Tamil & 61.70 & \textbf{64.08} (12) & 3.86 \\
\hline






\end{tabular}

}
\end{table}
\vspace{-3mm}

\vspace{-5mm}
\section{conclusion}
\label{sec: conclusion}
This study investigates the efficacy of acoustic measurements on speech intelligibility assessment, by using three dysarthric speech datasets of languages with different prosodic systems: English, Korean and Tamil. Acoustic measurements from voice quality, pronunciation, and prosody are analyzed. As multilingual analysis, examinations of the mean values of acoustic measurements and automatic intelligibility classification are conducted. This paper contributes to the automatic intelligibility classification of dysarthria by differentiating between language-independent and language-dependent measurements. PCC, PCV, and PCP are found to be independent of prosodic systems of language in intelligibility classification, validated by the two analyses. However, voice quality and prosody measurements generally present different aspects for different languages. Analysis of other datasets is also required to validate the generalizability of our findings. Cross-lingual experiments using language-independent features can be also explored in future work.



\section{Acknowledgment}
This work was supported by Institute of Information \& communications Technology Planning \& Evaluation (IITP) grant funded by the Korea government(MSIT) (No.2022-0-00223, Development of digital therapeutics to improve communication ability of autism spectrum disorder patients)




\bibliographystyle{IEEEbib}
\bibliography{refs}

\end{document}